\documentclass[11pt]{article}
\usepackage{graphicx}
\usepackage{amssymb}
\newcommand{\bsigma}{{\mbox{\boldmath $\sigma$}}}

\newcommand{\bm}{{{\mbox{\boldmath $m$}}}}
\newcommand{\bx}{{{\mbox{\boldmath $x$}}}}

\newcommand{\R}{\rm I \! R}

\begin{document}
\title{Loop corrections for message passing algorithms in
continuous variable models}

\author{Bastian Wemmenhove, Bert Kappen\\
Department of Biophysics,\\
SNN, Radboud University\\
Nijmegen, 6525 EZ Nijmegen, the Netherlands}


\maketitle

\begin{abstract}
In this paper we derive the equations for Loop Corrected Belief Propagation
on a continuous variable Gaussian model.  
Using the exactness of the averages for belief propagation for Gaussian
models, a 
different way of obtaining the covariances is found, 
based on Belief Propagation on cavity graphs. 
We discuss the relation of this 
loop correction algorithm to Expectation Propagation 
algorithms for the case in which the model is no longer 
Gaussian, but slightly perturbed by nonlinear terms. 
\end{abstract}
\section{Introduction}
Message passing techniques in graphical models allow for the computation
of (approximate) 
marginal probabilities in a time interval scaling polynomially in the 
model size. Their discovery has consequently revolutionized several 
fields of applications in the past years, of which error correcting codes
and vision are probably the most prominent examples. 
In many cases,
the corresponding graphs are loopy, implying either that the error
resulting from the
application of loopy belief propagation (BP) is negligible for the particular
model, or it 
can be tolerated for the particular purpose BP serves. In other cases 
more sophisticated refinements of BP are necessary, taking into account (part
of) the loop errors. 

Finding the optimal treatment of these ``loop errors'' 
motivates an active field of research, in which 
different solutions applying to different model classes are developed. 
For models involving many short loops, 
like on regular lattices, CVM type approaches
work well \cite{CVM}, or tree EP approaches \cite{TreeEP}. 
The latter may also be 
applied to correct for an incidental large loop. Unifying frameworks like the
Region graphs of \cite{RegionGraphs} 
lead to general strategies for selecting the basic clusters underlying
such approaches for general model classes.

A recent analysis has shown that the local update equations of BP may be
interpreted as the zero order term of an expansion in ``cavity connected
correlations''.
These quantities are parameterizations of the ``cavity distributions'', 
i.e., the 
distribution over neighbor variables of a central variable which has been
removed from the graph. The Bethe approximation and BP are recovered when this 
cavity distribution is assumed to factorize, whereas the first order
correction to the local update equations is obtained when one takes into
account the pair cumulants \cite{Rizzo}. 
Estimation of these pair cumulants is possible with extra runs of BP, allowing
for new polynomial time algorithms, reducing errors to order $1/N^{k+1}$ 
when applying algorithms of which running time scales with an extra factor 
of $N^k$ \cite{RizzoWem}.
Although this scaling seems heavy, the large benefit of the approach is
that it does not require selection of basic clusters or underlying 
tree-structures, since it takes into account the effect of all loops that
contribute to nontrivial correlations in the cavity distribution at once.

The above ``loop correction'' 
strategy is applicable in the class of models where a perturbative
expansion around the Bethe approximation makes sense, i.e., in models with
large loops and relatively weak interactions. The principal requirement 
is that the magnitude of pair
variable cumulants of cavity distributions is an order smaller than the 
single variable cumulants, and third order cumulants are even smaller, etc.
However, heuristics based on the strategy allow for other good algorithms 
performing well outside these parameter regimes \cite{Joris}.

So far the approach has been developed for discrete variable models on a
more abstract \cite{Rizzo,RizzoWem} versus practical level \cite{Joris}. 
In this
paper we apply the idea to graphical models for continuous variables.
We derive the loop corrected belief propagation equations 
for simple tractable Gaussian models, 
yielding a message passing scheme that, besides the correct average marginals,
also yields the correct variances. 
Besides that we discuss some approaches potentially 
applicable to cases in which extra function approximations are necessary, 
and the relation with expectation propagation.
A by-product of our loop corrected belief propagation equations is an algorithm
that calculates exact covariance matrices for Gaussian models like the one
discussed in \cite{WellingTeh}, but without explicitly using linear response.

\section{General ideas}
The error in the result of message passing techniques that are based on 
local approximations for variables that interact on a graph,
like belief propagation, may be viewed from two perspectives: 
\begin{itemize}
\item 
The error of the Bethe-approximation is due to the fact that loops in
the graph are neglected, such that nontrivial correlations between two
neighbors of one variable are neglected.
\item
The error is due to the fact that the functional parameterization of the
local marginals is not rich enough, such that it can at most be an 
approximation.
\end{itemize}
These two viewpoints may be argued to have the same meaning in the end, 
but nevertheless may lead to different strategies in the optimization of
the approximation, or the improvement of the results. If the second 
viewpoint is the starting point for algorithms like expectation propagation,
the first may be seen as the basic view for loop correction strategies.

Since expectation propagation applies well to continuous variable cases, but
loop correction schemes in the sense of \cite{Rizzo,Joris,RizzoWem} have not
been applied to continuous variable cases, it might be instructive to derive
corresponding equations and compare them to expectation propagation
approaches.

With this motivation in mind, we will
firstly analyze a loop correction scheme to BP in Gaussian models. Given
this scheme, we will discuss possible generalizations suitable for cases
in which the model is no longer tractable.
\subsection{Model introduction}
The model which we will initially consider is a Gaussian model of
$N$ interacting variables, denoted by ${\bsigma} \in \R^N$
of which the total probability distribution is given by
\begin{eqnarray}
P(\bsigma) & = & Z^{-1}\prod_{i=1}^N \psi_i(\sigma_i)
\prod_{j<k}^N \psi_{jk}(\sigma_j,\sigma_k)\nonumber \\
\psi_i(\sigma_i) & = & \exp \left[ -\frac{1}{2s_i}(\sigma_i - \mu_i)^2 \right]
\nonumber \\
\psi_{jk}(\sigma_j,\sigma_k)& = &\exp\left[J_{jk}\sigma_j\sigma_k\right]
\end{eqnarray}
thus the variables $i$ have their own Gaussian local potential with average
$\mu_i$ and variance $s_i$, but interact in a pairwise manner with 
variables $j$ via the
interaction $J_{ij}$. Obviously, 
$Z = \int d{\bsigma} \prod_{i=1}^N \psi_i(\sigma_i)
\prod_{j<k}^N \psi_{jk}(\sigma_j,\sigma_k)$. We will denote the 
neighborhood of variable $i$ on the graph by $\partial_i$, i.e.
$\partial_i = \{ j | J_{ij} \neq 0 \}$.
\subsection{The ``cavity equations''}
\label{basicideas}
The following analysis will be based on the loop correction equations of 
\cite{Rizzo}, which were applied to discrete binary variables. 
The current generalization to continuous variables is a straightforward
application of these ideas.
We write down an expression for the joint probability 
$P^{(ij)}(\sigma_i,\sigma_j)$ of variables $\sigma_i$ and $\sigma_j$
on the model from which the interaction $J_{ij}$ has been removed in two
different ways. The first is in terms
of the {\em cavity distribution} of variable $i$, 
$P^{(i)}(\sigma_{\partial_i})$, i.e., the
joint distribution over the neighbors of $i$ in the model from which 
$i$ has been removed, and the second in terms of the cavity distribution
of variable $j$, 
$P^{(j)}(\sigma_{\partial_j})$:
\begin{eqnarray}
P^{(ij)}(\sigma_i,\sigma_j)&  = &\frac{1}{Z_1}
\int {\rm d} \bsigma_{\partial_i \setminus j} P^{(i)}(\bsigma_{\partial_i})
\exp\left[-(\sigma_i - \mu_i)^2/(2s_i) 
+\sigma_i \sum_{l\in \partial i \setminus j}
 J_{il}\sigma_l\right] \\
P^{(ij)}(\sigma_i,\sigma_j)&  = &\frac{1}{Z_2}
\int {\rm d} \bsigma_{\partial_j \setminus i} P^{(j)}(\bsigma_{\partial_j})
\exp \left[-(\sigma_j - \mu_j)^2/(2s_j)
+\sigma_j \sum_{l\in \partial_j \setminus i}
 J_{jl}\sigma_l \right]
\end{eqnarray}
With respect to this marginal distribution two ways of writing the
moment 
\begin{equation}
\langle \sigma_i \rangle^{(ij)} \equiv \int {\rm d} \sigma_i \sigma_i
P^{(ij)}(\sigma_i,\sigma_j)
\end{equation} 
are 
\begin{eqnarray}
\langle\sigma_i \rangle^{(ij)}& =&\frac{1}{Z_1}
\int {\rm d} \bsigma_{\partial i} {\rm d}
\sigma_i  P^{(i)}(\bsigma_{\partial i}) \sigma_i
\exp \left[-(\sigma_i - \mu_i)^2/(2s_i)
+\sigma_i \sum_{l\in \partial i \setminus j}
 J_{il}\sigma_l \right] \nonumber \\
\langle \sigma_i \rangle^{(ij)} &=& \frac{1}{Z_2}
\int {\rm d} \bsigma_{\partial j} {\rm d}
\sigma_j  P^{(j)}(\bsigma_{\partial j}) \sigma_i
\exp \left[-(\sigma_j - \mu_j)^2/(2s_j)
+\sigma_j \sum_{l\in \partial j \setminus i}
 J_{jl}\sigma_l\right]
\label{momi}
\end{eqnarray}
which may be written in terms of effective
measures
\begin{eqnarray}
\langle f(\bsigma_{\partial_i}) \rangle_{i\to j} &\equiv& Z_{i\to j}^{-1} 
\int {\rm d} 
\bsigma_{\partial i} f(\bsigma_{\partial_i}) P^{(i)}(\bsigma_{\partial i})
\exp\left[ \mu_i \sum_{l\in \partial i \setminus j}J_{il}\sigma_l
+\frac{s_i}{2}
\sum_{l,k \in \partial i \setminus j}J_{il}J_{ik} \sigma_l \sigma_k\right]
\nonumber \\
\langle f(\bsigma_{\partial j}) \rangle_{j\to i} &\equiv& 
Z_{j\to i}^{-1}  \int {\rm d} \bsigma_{\partial j}  f(\bsigma_{\partial j})
P^{(j)}(\bsigma_{\partial j})
\exp\left[\mu_j \sum_{l\in \partial j \setminus i}J_{jl}\sigma_l 
+\frac{s_j}{2}
\sum_{l,k \in \partial j \setminus i}J_{jl}J_{jk} \sigma_l \sigma_k\right]
\nonumber \\
\end{eqnarray}
where $Z_{i\to j}$ and $Z_{j\to i}$ are the corresponding normalization 
constants.
In terms of these measures, the equations (\ref{momi}) lead to
\begin{eqnarray} 
\langle \sigma_i \rangle_{j \to i} = \mu_i +s_i
\sum_{l \in \partial i \setminus j}J_{il}\langle \sigma_l \rangle_{i \to j}
\label{firstmom}
\end{eqnarray}
The above procedure may be repeated for all other moments of the distribution
$P^{(ij)}(\sigma_i)$, e.g.
\begin{eqnarray}
\langle \sigma_i^2 \rangle_{j \to i}& =& s_i + \mu_i^2 + 2s_i\mu_i\sum_{l \in
\partial i \setminus j} J_{il} \langle \sigma_l \rangle_{i \to j}
+ s_i^2
\sum_{l,k \in \partial i \setminus j} J_{il}J_{ik} \langle \sigma_l \sigma_k 
\rangle_{i \to j}
\label{secondmom}
\end{eqnarray}
etc.
The moments of the true marginal distributions are integrals with
respect to different measures, e.g.:
\begin{equation}
\langle \sigma_i \rangle 
= \mu_i + s_i\sum_{l \in \partial i}J_{il} \langle \sigma_l \rangle_i
\label{average}
\end{equation}
with $\langle f({\bsigma})\rangle = \int d{\bsigma} P({\bsigma}) 
f({\bsigma})$ and
\begin{eqnarray}
\langle f(\bsigma_{\partial_i}) \rangle_i &\equiv& Z_i^{-1}
\int {\rm d} \bsigma_{\partial i} f(\bsigma_{\partial_i})  
P^{(i)}(\bsigma_{\partial i}) \exp\left[
\mu_i \sum_{l\in \partial i}J_{il}\sigma_l 
+ \frac{s_i}{2}
\sum_{l,k \in \partial i}J_{il}J_{ik} \sigma_l \sigma_k \right]
\label{measurei}
\end{eqnarray}
All these measures reduce to functions of the above
mentioned cavity distributions,
which are the unknown functions of interest.
It is clear however, that so far we have not specified enough local 
equations to solve for the full
cavity distributions $P^{(i)}(\bsigma_{\partial_i})$. 
If we restrict ourselves, for the moment, to Gaussian models, we will
be able to perform the integrations and find exact local message
passing equations.
We note that for a more general
type of model, such local computations will be insufficient, but may
be used as a basis for an approximation when an appropriate set of
approximating functions is chosen, characterized by a finite set of parameters.

\section{Gaussian cavity distributions}
Notice that the belief propagation is recovered when one chooses to 
approximate the cavity distribution by 
a factorizing one, i.e. $P^{(i)}(\bsigma_{\partial_i})\sim 
\prod_{j\in \partial_i} Q^{(i)}(\sigma_j)$. This parameterization
includes the exact result when the graph
is a tree, since then there can be no nontrivial correlations between 
variables in any cavity set $\partial_i$ when $i$ is absent. When there
are loops in the graph, corrections to this parameterization are desirable.
Various parameterizations of these corrections are possible in principle,
and in \cite{Rizzo} it was suggested to expand the cavity distributions
in the cumulants, an expansion that is appropriate when either interactions
are weak or loops are long.
 
For a Gaussian model, the cavity distributions are completely specified by
their averages and covariances, such that including the second order cumulants
(the first order correction to belief propagation) yields exact equations.
In the following we investigate the structure of the corresponding 
equations and identify the exact correction to Gaussian belief propagation.
An appropriate (and exact)
parameterization of the cavity distribution is 
\begin{equation}
P^{(i)}(\bsigma_{\partial i}) \sim \exp\left[-\frac{1}{2}(\mathbf{\bsigma}_{
\partial_i}-
\mathbf{m}^i)^T[D_i + A_i]^{-1}(\mathbf{\bsigma}_{\partial_i}-\mathbf{m}^i)
\right]
\label{gausscav}
\end{equation}
where we have decomposed the covariance
matrix in a diagonal part ($D_i$) and an
off-diagonal part ($A_i$), both having the dimensions of the cavity set. 
The Bethe
approximation, for which cavity distributions factorize, corresponds to
neglecting the off-diagonal components $A_i$.
The matrices $D_i$ and vectors $\mathbf{m}^i$
are found through consistency equations.
In the following we will denote the vector $\mathbf{J}_i$ (again the
dimensions of vectors $\mathbf{m}^i$ and $\mathbf{J}_i$ are 
equal to that of the cavity set, $|\partial_i|$) for which 
$J_{ij}=0$ as $\mathbf{J}_i^j$.
The consistency equations (\ref{firstmom}), by Gaussian integration, are found
to be
\begin{eqnarray}
&& \hspace*{-10mm}
\left\{[(D_j+ A_j)^{-1} -s_j\mathbf{J}_j^i\mathbf{J}_j^{iT}]^{-1}
[(D_j+ A_j)^{-1}\mathbf{m}^j +  \mu_j \mathbf{J}_j^i]
\right\}_i \nonumber \\
&=& \mu_i + s_i\mathbf{J}_i^{jT}
\left\{[(D_i+ A_i)^{-1} -s_i\mathbf{J}_i^j\mathbf{J}_i^{jT}]^{-1} 
[(D_i+ A_i)^{-1}\mathbf{m}^i + \mu_i
\mathbf{J}_i^j]\right\}
\label{updateX}
\end{eqnarray}
From the relations of the
variances, equation (\ref{secondmom}), we find:
\begin{eqnarray}
\{[(D_j+ A_j)^{-1} -s_j\mathbf{J}_j^i\mathbf{J}_j^{iT}]^{-1}\}_{ii}
&=& s_i+ s_i^2 \mathbf{J}_{i}^{jT}
[(D_i+ A_i)^{-1} -s_i\mathbf{J}_i^j\mathbf{J}_i^{jT}]^{-1}
\mathbf{J}_i^j
\label{updateD}
\end{eqnarray}
For each cavity distribution $D_j$ and $\mathbf{m}^j$ the number of
pairs of equations is equal to the number of variables in the cavity set.
Thus, given a covariance matrix $A$, the diagonals $D$ can be determined with
the second equation, and subsequently the average values $\mathbf{m}$ 
can be determined with the first equation. The marginal distributions
then follow directly, since all variables are now known.
Substituting (\ref{gausscav}) into (\ref{average}),
we find
\begin{eqnarray}
\langle \sigma_i \rangle
&=& \mu_i + s_i \mathbf{J}_{i}
\left[(D_i+ A_i)^{-1} -s_i\mathbf{J}_i\mathbf{J}_i^T\right]^{-1} 
\left[(D_i+ A_i)^{-1}\mathbf{m}^i + \mu_i \mathbf{J}_i\right]
\end{eqnarray}
and for the second moment
\begin{eqnarray}
\langle \sigma_i^2 \rangle
&=& \langle \sigma_i \rangle^2 + s_i\left\{1+s_i\mathbf{J}_{i}^T
[(D_i+A_i)^{-1} - s_i \mathbf{J}_i \mathbf{J}_i^T ]^{-1} \mathbf{J}_i\right\}
\end{eqnarray}
The only obstacle in solving these {\it exact} equations 
is yet obtaining the off-diagonal covariances $A_i$ for
each cavity set $\partial_i$. 

Simply neglecting them, setting $A_i = 0$, we 
should recover the BP equations for the Gaussian model. 

Using response propagation it is possible to estimate the 
covariances, which leads to an improvement in the results when they 
are small for the binary case \cite{Rizzo,RizzoWem}.
In the Gaussian case, where results from response propagation are exact,
\cite{WellingTeh,WeissJordan}, this procedure should thus yield exact results
provided response propagation and belief propagation both converge.

\section{Loop corrected belief propagation}
Using the identity
\begin{eqnarray}
[A + XBX^T]^{-1} &=&
A^{-1}- A^{-1}X(B^{-1}+X^T
A^{-1}X)^{-1}X^TA^{-1}
\end{eqnarray}
we may write
\begin{eqnarray}
[(D_i + A_i)^{-1}-s_i\mathbf{J}_i^j\mathbf{J}_i^{jT}]^{-1} & = &D_i+A_i 
+ \frac{(D_i+A_i)
\mathbf{J}_i^j \mathbf{J}_i^{jT}(D_i+A_i)}{1/s_i-\mathbf{J}_i^{jT}
(D_i+A_i)\mathbf{J}_i^j}
\end{eqnarray}
Defining
\begin{eqnarray}
\alpha_i^j & \equiv &\mathbf{J}_i^{jT}(D_i+A_i)\mathbf{J}_i^j \\
\alpha_i &\equiv &\mathbf{J}_i^{T}(D_i+A_i)\mathbf{J}_i \\
\epsilon_j^i &\equiv &[(D_j+A_j)\mathbf{J}_j^i]_i = [A_j \mathbf{J}_j^i]_i
\end{eqnarray}
and writing $v_k^i$ for the (diagonal) entries of $D_i$ where $k$ runs over
$\partial_i$,
we find that equation (\ref{updateD}) yields
\begin{eqnarray}
v_i^j + \frac{s_j}{1-s_j\alpha_j^i}(\epsilon_j^i)^2 = \frac{s_i}{1-s_i\alpha_i^j}
\end{eqnarray}
After similar simplification of equation (\ref{updateX}), the updates for the 
message variances and averages become
\begin{eqnarray}
v_i^j  &=& 
\frac{s_i}{1-s_i\alpha_i^j}- \frac{s_j}{1-s_j\alpha_j^i}(\epsilon_j^i)^2 \\
m_i^j &=& \frac{s_i}{1-s_i\alpha_i^j}\left[\frac{\mu_i}{s_i}+
\sum_{l\in \partial i \setminus j}J_{il}m_l^i\right]
- \frac{s_j \epsilon_j^i}{1-s_j\alpha_j^i}\left[ \frac{\mu_j}{s_j}
+ \sum_{l \in \partial j \setminus i}J_{jl}m_l^j\right]
\label{messages}
\end{eqnarray}
and the final marginals are given by
\begin{eqnarray}
v_i & = &\frac{s_i}{1-s_i\alpha_i} \\
m_i & = & v_i\left[ \frac{\mu_i}{s_i} + \sum_{l\in \partial i} J_{il}m_l^i
\right]
\label{exactmargs}
\end{eqnarray}
Indeed the BP equations follow for $A_i=0$, since in that case 
$\epsilon_i^j=0$, $\alpha_i^j = \sum_{k \in \partial_i \setminus j} J_{ik}^2
v_k^i$ and
$\alpha_i = \sum_{j\in \partial_i} J_{ij}^2v_j^i$, such that the equations
(modulo a transformation) reduce to the ones in \cite{WeissJordan}.

The above equations allow one to explicitly interpret the meaning of
the belief propagation messages, and write down expressions for their error.
Indeed the messages in equation (\ref{messages}) represent averages and 
variances of 
cavity distributions, i.e., of the model in absence of a variable. 
An interesting side result in this respect comes from the observation in 
\cite{WeissJordan}
that the averages calculated via belief propagation are exact when the 
algorithm converges. It follows that the message $m_i^j$ calculated via
equation (\ref{messages}) is equal to $m_i$ on a graph from which variable
$j$ is removed calculated via ordinary belief propagation. In the next
section we use this observation, together with similar arguments, to
obtain some more exact results from ordinary belief propagation variables
alone.

\section{An alternative way to calculate the error in $v_i$ for Gaussian 
models}
The form of the loop corrected belief propagation
equations imposes a relation between the BP errors in $v_i$
and the messages $m_l^i$ for Gaussian models. 
Comparing the result of equations (\ref{exactmargs}) with and without
cavity covariances, one may show that
\begin{eqnarray}
m_i^{\rm LC}  & = & m_i^{\rm BP} + v_i^{\rm LC }
\Biggl(m_i^{\rm BP}[
\alpha_i^{\rm LC}-\alpha_i^{\rm BP}] + 
\sum_{l \in \partial_i} J_{il} [m_l^{i \ {\rm LC}}
- m_l^{i \ {\rm BP}}]\Biggr)
\end{eqnarray}
Now, since the BP averages are exact whenever BP converges \cite{WeissJordan},
\begin{eqnarray}
m_i^{\rm BP}[\alpha_i^{\rm LC}-\alpha_i^{\rm BP}]
= -\sum_{l\in \partial i}J_{il}[m_l^{i \ {\rm LC}}
- m_l^{i \ {\rm BP}}]
\end{eqnarray}
Due to the interpretation of the message averages as cavity parameters, we have
access to $m_l^{i\ {\rm LC}}$, since 
\begin{eqnarray}
m_l^{i {\rm \ LC}} = m^{(i)\ {\rm BP}}_l
\end{eqnarray}
i.e. this is the average of variable $l$ on the graph without $i$, which 
may be obtained by running BP on the graph without variable $i$.
Thus by running BP on the original graph once and running it on the graph 
without $i$, we can calculate $v_i^{\rm \ LC}$ by using equation
(\ref{exactmargs}) and writing
\begin{eqnarray}
v_i^{\rm LC} = \frac{s_i}{1-s_i
\left[\alpha_i^{\rm BP}+\left(m_i^{\rm BP}\right)^{-1}
\sum_{l\in \partial i}J_{il}[m_l^{(i) \ {\rm BP}}
- m_l^{i \ {\rm BP}}\right]}
\end{eqnarray}
provided that $m_i^{\rm BP}\neq 0$.
Similar considerations (see appendix) lead to a procedure for calculating
the entire covariance matrix using BP: condensing notation
\begin{eqnarray}
\kappa_j^i & \equiv &  
J_{ij}v_j^{i{\rm BP}}-\frac{[m_j^{(i){\rm BP}}-m_j^{i\ {\rm BP}}]}
{m_i^{\rm BP}}
\\
u_j^i &\equiv &m_j^{(i){\rm BP}} + m_i^{\rm BP}\kappa_j^i \\
v_i &\equiv &  \frac{s_i}{1-s_i
\left[\alpha_i^{\rm BP}+\left(m_i^{\rm BP}\right)^{-1}
\sum_{l\in \partial i}J_{il}[m_l^{(i) \ {\rm BP}}
- m_l^{i \ {\rm BP}}\right]}
 \\
m_i &\equiv& m_i^{\rm BP}
\end{eqnarray}
we have the following equations:
\begin{eqnarray}
\langle \sigma_i^2 \rangle & = & v_i + m_i^2 \\
\langle \sigma_i \sigma_j \rangle & = &m_i u_j^i + v_i \kappa_j^i \\
\langle \sigma_j \sigma_k \rangle & = & u_j^i u_k^i + v_i \kappa_j^i \kappa_k^i
+ \langle \sigma_j\sigma_k\rangle^{(i)}
\label{growcov}
\end{eqnarray}
These equations suggest inverting matrices by calculating correlation matrices
on growing graphs might be a useful application. By subsequently attaching
new variables to the graph and running BP, one finds the full correlation
matrix with $N$ runs of BP, just as with the procedure described in 
\cite{WellingTeh}, but the cost of the BP runs is halved since the graph is 
growing along with the BP runs.
However, we should not overlook the fact that the equations above introduce 
large number of additions and multiplications, such that in the end 
the total computational complexity for inverting a sparse matrix 
is similar to other well-known methods.

\section{Nonlinear models: connections with EP}
The fact that loop corrections in the above form are able to correct for the
total BP error in the (co)variances is of course due to the Gaussian nature
of the model. In discrete models, exact parameterizations of the full 
distribution by use of local marginals only is in general not possible, but
loop corrections are able to increase the accuracy of the Bethe approximation.
Thus the above formalism might
seem a promising basis for extensions to models that are not exactly 
tractable, possibly as an alternative for related algorithms like
Expectation Propagation (EP) \cite{Minka2}. 
Since BP may be viewed as a special case of EP, we may hope for some
generalizations of loop corrections
equations, with some relation to EP, applicable in cases where function 
approximations become necessary. 
However, the specific form of EP equations
very much depends on the choice of the approximating family of functions one
chooses. The equivalence with BP corresponds to a family of approximate EP
functions that fully factorizes over the variables of the model 
\cite{Minka2}. For loop corrected BP strategies, we expect a relationship
with EP approaches based on larger local neighborhoods. 


We will investigate the relation to EP by deriving
equations for models with general nonlinear single-variable potentials, 
i.e.,
$\psi_i(\sigma_i) \to \psi_i(\sigma_i)e^{-V_i(\sigma_i)}$,
as one might expect in vision problems with nonlinear observation functions.

\subsection{Full Gaussian EP}
The Gaussian loop corrections approach seems rather similar to an EP 
approach where one 
includes a full Gaussian in the approximating target distribution.
The standard EP formalism for this approach is to choose as an 
approximate distribution
\begin{eqnarray}
q(\bx) & \sim & \exp\left[ -\frac{1}{2}(\bx - \bm)^T \Sigma^{-1}
(\bx -\bm) \right]
\end{eqnarray}
where
\begin{eqnarray}
\Sigma^{-1}& = &\Sigma_{\rm g}^{-1} + \sum_i (\Sigma^i)^{-1} \\
\Sigma^{-1}\bm & = & 
\Sigma_{\rm g}^{-1}\bm_{\rm g} + \sum_i(\Sigma^i)^{-1}\bm^i
\label{fullm}
\end{eqnarray}
and the subscript g stands for Gaussian, as it represents the Gaussian
contribution to the full joint probability. The remaining terms relate to
approximations of the single node potentials in the following manner
\begin{eqnarray}
q(\bx) = q_g(\bx)\prod_i \overline{f}^i(\bx)
\end{eqnarray}
Here again $q_g(\bx)=\exp[-(\bx-\bm_g)^T\Sigma_g^{-1}(\bx-\bm_g)/2]$
and $\overline{f}^i(\bx)$ is the standard Minka notation \cite{Minka2}
for a term that
approximates an intractable contribution, in our case
\begin{eqnarray}
f^i(\bx) = f^i(x_i)= e^{-V_i(x_i)}
\end{eqnarray}
Updating the parameters $\bm^i$ and $\Sigma^i$ proceeds in the usual way:
first, for a term $i$, the contribution of its 
approximation is removed from the full joint:
\begin{eqnarray}
q^{\setminus i}(\bx) \sim \frac{q(\bx)}{\overline{f}^i(\bx)}
\end{eqnarray}
meaning that
\begin{eqnarray}
(\Sigma^{\setminus i})^{-1} & = &\Sigma_{\rm g}^{-1} + \sum_{j(\neq i)} 
(\Sigma^j)^{-1} \\
(\Sigma^{\setminus i})^{-1}\bm^{\setminus i} & = & 
\Sigma_{\rm g}^{-1}\bm_{\rm g} + \sum_{j(\neq i)}(\Sigma^j)^{-1}\bm^j
\end{eqnarray}
Then the new value of the full parameters is obtained by defining
\begin{eqnarray}
\hat{p}(\bx)& = &\frac{q^{\setminus i}(\bx)f^i(x_i)}{
\int d\bx q^{\setminus i}(\bx)f^i(x_i)} 
\end{eqnarray}
and minimizing
\begin{eqnarray}
KL(\hat{p}|q) & = & \int d\bx \hat{p}(\bx)\log \left[ \frac{\hat{p}(\bx)}
{q(\bx)}
\right] \nonumber \\
& \sim & \int d\bx q^{\setminus i}(\bx)f^i(x_i) \log
\left[ \frac{ q^{\setminus i}(\bx)f^i(x_i)}{q_1(\bx)
q_2(x_i)}\right] \nonumber \\
& = & \int d\bx q^{\setminus i}(\bx)f^i(x_i) \left\{ \log
\left[  \frac{ f^i(x_i)}{
q_2(x_i)}\right] + \log\left[ \frac{ q^{\setminus i}(\bx)}{q_1(\bx)
}\right] \right\}
\end{eqnarray}
where we have taken the liberty of decomposing the Gaussian function 
$q(\bx)$ into a Gaussian that depends only on $x_i$ and a remaining
Gaussian depending on the whole vector $\bx$. Since both $q_1(\bx)$ and
$q^{\setminus i}(\bx)$ are Gaussians, the KL-divergence is minimal when 
they are equal and thus we have to minimize
\begin{eqnarray}
KL(\hat{p}|q) & = & \int dx_i q^{\setminus i}(x_i) f^i(x_i) \log
\frac{f^i(x_i)}{q_2(x_i)}
\end{eqnarray}
with respect to $q_2(x_i)$, 
where $q^{\setminus i}(x_i) = \int d\bx_{\setminus x_i} q^{\setminus i}(\bx)$
and it is clear that $q_2(x_i)$ is parameterized by $(\Sigma^i)^{-1}$ and 
$m^i$, the only parameters to be updated. We furthermore deduce that these
parameters contribute only to single entries in the matrices and vectors
(i.e. they are scalars). Thus
\begin{eqnarray}
m^i &=& Z^{-1}
\int dx_i ~ x_i ~ q^{\setminus i}(x_i)f^i(x_i)\\
\Sigma^i & = & Z^{-1}
\int dx_i ~x_i^2 ~q^{\setminus i}(x_i) f^i(x_i) - (m^i)^2
\\
 Z &= &
\int dx_i  ~ q^{\setminus i}(x_i) f^i(x_i)
\label{updates}
\end{eqnarray}
The marginalization of 
$q^{\setminus i}(\bx)$ yields,
\begin{equation}
q^{\setminus i}(x_i) \sim \exp\left[ -\frac{(x_i-m_i^{\setminus i})^2}
{2 \Sigma_{ii}^{\setminus i}}
\right]
\end{equation}
where
\begin{eqnarray}
\label{iters}
\Sigma^{\setminus i}& = &\left[ (\Sigma_{\rm g})^{-1} + {\rm diag}_{
\setminus i}\left(\frac{1}{\Sigma^j}\right)\right]^{-1} \\
m^{\setminus i}_i & = & \sum_l (\Sigma^{\setminus i})_{il}\left[
[(\Sigma_{\rm g}^{-1})\bm_{\rm g}]_l + \frac{m^l}{\Sigma^l}(1-\delta_{il})
\right]
\end{eqnarray}
Thus the most costly computations are the inversion in equation (\ref{iters})
and the one-dimensional integrations of (\ref{updates}). For very large models
the inversions may become prohibitive, but otherwise this scheme seems 
efficient, since the ``cavity covariances'' do not have to be 
computed separately but are implicitly present in these inversions, and are 
optimal with respect to the minimization of the KL-divergence.

\subsection{Loop corrections formulation}
In this subsection we will discuss a possible generalization of the 
loop correction scheme for the model discussed in the previous subsection.
The same model with nonlinear single-variable potentials may be 
tackled starting from the loop correction scheme at the beginning of this
paper, by slightly generalizing it to the case where
$\psi_i(\sigma_i) \to \psi_i(\sigma_i)e^{-V_i(\sigma_i)}$. 
The formalism at the beginning of this paper may still be applied when
the Gaussian parameterization of the distributions 
$P^{(i)}(\bsigma_{\partial_i})$ for all $i$ is an 
approximation.
For given estimates of the
covariance matrices $A_i$ we then find:
\begin{eqnarray}
m_i^j & = & \frac{\int d\sigma \sigma \exp \left\{ 
-\Phi_i^j(\sigma,\mathbf{m}^i,\alpha_i^j(A_i,\{v_l^i\}))
\right\}}{\int d\sigma \exp\left\{ 
-\Phi_i^j(\sigma,\mathbf{m}^i,\alpha_i^j(A_i,\{v_l^i\}))
\right\}}
\nonumber \\
&&-\frac{\epsilon_j^i \int d\tau \tau \exp \left\{ 
-\Phi_j^i(\tau,\mathbf{m}^j,\alpha_j^i(A_i,\{v_l^i\})) \right\}}{
\int d\tau \exp \left\{ 
-\Phi_j^i(\tau,\mathbf{m}^j,\alpha_j^i(A_i,\{v_l^i\})) \right\}} \\
v_i^j & = & \frac{\int d\sigma \sigma^2  \exp \left\{ 
-\Phi_i^j(\sigma,\mathbf{m}^i,\alpha_i^j(A_i,\{v_l^i\}))
\right\}}{\int d\sigma  \exp \left\{ 
-\Phi_i^j(\sigma,\mathbf{m}^i,\alpha_i^j(A_i,\{v_l^i\}))
\right\}}
\nonumber \\
&&-\frac{\int d\tau (m_i^j+\epsilon_j^i\tau)^2
 \exp \left\{ -\Phi_j^i(\tau,\mathbf{m}^j,\alpha_j^i(A_i,\{v_l^i\}))\right\}}{
\int d\tau  \exp \left\{ 
-\Phi_j^i(\tau,\mathbf{m}^j,\alpha_j^i(A_i,\{v_l^i\})) \right\}} \\
m_i  & = & \frac{\int d\sigma \sigma \exp \left\{ 
-\Phi_i(\sigma,\mathbf{m}^i,\alpha_i(A_i,\{v_l^i\}))
\right\}}{\int d\sigma \exp\left\{ 
-\Phi_i(\sigma,\mathbf{m}^i,\alpha_i(A_i,\{v_l^i\}))
\right\}} \\
v_i + m_i^2 & = & \frac{\int d\sigma \sigma^2 \exp \left\{ 
-\Phi_i(\sigma,\mathbf{m}^i,\alpha_i(A_i,\{v_l^i\}))
\right\}}{\int d\sigma \exp\left\{ 
-\Phi_i(\sigma,\mathbf{m}^i,\alpha_i(A_i,\{v_l^i\}))
\right\}}
\label{genupdates}
\end{eqnarray}
with
\begin{eqnarray}
\Phi_i^j(\sigma,\mathbf{m}^i,\alpha_i^j(A_i,\{v_l^i\}))& = &
\frac{(\sigma-\hat{m}_i^j)^2}{2\hat{v}_i^j}+V_i(\sigma) \\
\hat{v}_i^j & = &\frac{s_i}{1-s_i\alpha_i^j(A_i,\{v_l^i\})}\\
\hat{m}_i^j &=& \hat{v}_i^j\left[\frac{\mu_i}{s_i} + \sum_{k\in \partial i
\setminus j}J_{ik}m_k^i\right]\\
\Phi_i(\sigma,\mathbf{m}^i,\alpha_i(A_i,\{v_l^i\})) & = & 
\frac{(\sigma-\hat{m}_i)^2}{2\hat{v}_i}+V_i(\sigma) \\
\hat{v}_i  & = &\frac{s_i}{1-s_i\alpha_i(A_i,\{v_l^i\})}\\
\hat{m}_i &=& \hat{v}_i\left[\frac{\mu_i}{s_i} + \sum_{k\in \partial i
}J_{ik}m_k^i\right]
\label{genphi}
\end{eqnarray}
On the one hand, it is easy to check that these equations reduce to
the BP equations with loop correction when $V_i(\sigma_i)=0$ for all $i$, 
i.e. $m_i^j = \hat{m}_i^j -\epsilon_j^i \hat{m}_j^i$ and 
$v_i^j = \hat{v}_i^j - (\epsilon_j^i)^2 \hat{v}_j^i$. In that case they
should be equivalent to the full Gaussian EP approach of the previous
subsection as well, since both treatments are exact in this limit.
On the other hand, when we take $V_i(\sigma_i) \neq 0$ and
$A_i=0$ for all $i$,
these updates are somewhat similar to EP with completely factorizing
Gaussian approximate target distribution (i.e., deriving
equations starting from diagonal $\Sigma$).
The slight difference is due to the fact that the propagated expectations
$m_i^j$ and $v_i^j$ parameterize approximate cavity distributions 
(i.e. in absence of one neighboring variable) and not the actual target
marginal distributions. Thus the KL-divergence with an approximate 
factorizing cavity distribution is minimized and not with the approximate
target distribution.
When $V_i(\sigma_i)=0$ for all $i$, this difference 
vanishes, and the algorithm reduces to 
EP with a factorizing Gaussian
as approximate joint distribution, which, in that limit (where integrals
may be performed exactly) is equivalent to ordinary BP.

When optimizing the approximations for marginal moments is the objective of 
the algorithm, the approach of this subsection is obviously not optimal, 
since instead
moments of cavity distributions are optimized in the integrals that calculate
the messages.

\subsection{Alternative loop correction formalism}
Inspired by the above observations regarding the optimization of the 
marginal moments of the target approximation, one may derive 
alternative consistency equations as in \cite{Joris}, starting from the 
expressions for the actual marginals, such that the integrations include
full sets of neighboring factors. Once again, we approximate the cavity 
distributions by Gaussians, and find
\begin{eqnarray}
m_i^j &= &\langle \sigma_i \rangle_{\hat{i}} - [J_{ij}v_i^j + \epsilon_j^i]
\langle \sigma_j \rangle_{\hat{j}} \\
v_i^j &=& \left[ \langle \sigma_i^2 \rangle_{\hat{i}} -
\left(\langle \sigma_i \rangle_{\hat{i}} \right)^2 \right] -
\left( J_{ij}v_i^j + \epsilon_j^i\right)^2\left[
\langle \sigma_j^2 \rangle_{\hat{j}} - \left(\langle \sigma_j
\rangle_{\hat{j}}\right)^2 \right] 
\end{eqnarray}
with
\begin{eqnarray}
\langle \sigma_i \rangle_{\hat{i}}  & = &
\frac{\int d\sigma_i \sigma_i \exp\left[ 
-\Phi_i(\sigma_i,\mathbf{m}^i,\alpha_i(A_i,\{v_l^i\}))
\right]}
{\int d\sigma_i \exp\left[ -\Phi_i(\sigma_i,\mathbf{m}^i,\alpha_i(A_i,\{v_l^i\}))
\right]} \qquad \\
\langle \sigma_i^2 \rangle_{\hat{i}}  & = &
\frac{\int d\sigma_i \sigma_i^2 
\exp\left[ -\Phi_i(\sigma_i,\mathbf{m}^i,\alpha_i(A_i,\{v_l^i\}))
\right]}
{\int d\sigma_i \exp\left[ -\Phi_i(\sigma_i,\mathbf{m}^i,\alpha_i(A_i,\{v_l^i\}))
\right]} \qquad
\end{eqnarray}
For $A_i=0$ this reduces to EP with fully factorizing Gaussian, and again
$V_i(\sigma_i) = 0$ leads to equations which are equivalent to BP. 
A suitable choice of $A_i$ should make the fixed points
of the above equations equivalent to the full-Gaussian EP equations at the
beginning of this section, since both approaches optimize the marginal moments
of each variable, given a Gaussian interaction matrix with the rest of the 
model. However, the benefit of full Gaussian EP is that this Gaussian 
interaction matrix is optimized on the way, albeit at the cost of an 
inversion at each iteration, while the loop corrected approach desires an 
estimate of $A_i$ as input, which is not further updated. 

Thus loop corrections are an alternative for the current type of
model only if these inversions are so costly that approximations of the above
form are sensible.

\subsection{Estimating $A_i$: response propagation}
In the above formalism, an approximation for the cavity
covariance matrix $A_i$ may be obtained by applying a linear response
algorithm to the graph from which variable $i$ has been removed. The entries
of $A_i$ for variables $j,k \in \partial_i$ follow from
\begin{eqnarray}
\langle \sigma_j \sigma_k\rangle^{(i)} -\langle
\sigma_j \rangle^{(i)}\langle \sigma_k
\rangle^{(i)}&=& s_k \frac{\partial \langle
\sigma_j \rangle^{(i)}}{\partial \mu_k} \\
& = & s_k \frac{\partial m_j^{(i)}}{\partial \mu_k}
\end{eqnarray}
Thus derivatives of average messages should be computed. Using 
(\ref{genupdates}) and (\ref{genphi}), we estimate them by
taking the derivatives of these update equations neglecting the corresponding
cavity covariances (setting $A_j=0$ for all $j\neq i$):
\begin{eqnarray}
\frac{\partial m_j^{(i)}}{\partial \mu_k} & = & 
v_j^{(i)}\left[ \frac{\delta_{jk}}{s_j}+\sum_{l\in \partial_j} J_{jl}
\frac{\partial m_l^{j(i)}}{\partial \mu_k}\right]\\
\frac{\partial m_l^{j(i)}}{\partial \mu_k} & = & 
v_l^{j(i)}\left[ \frac{\delta_{lk}}{s_l}+\sum_{n\in \partial_l \setminus
j} J_{ln}
\frac{\partial m_n^{l(i)}}{\partial \mu_k}\right]
\end{eqnarray}
The reasoning here is that an estimate of the covariance matrix in 
``zeroth order'' enables a first order corrected version (see reference
\cite{Rizzo}) of the expectation
propagation algorithm, equations (\ref{genupdates}) and (\ref{genphi})
with $A_i=0 \  \forall i$.
Note that, given the values of the single node variances, which 
have to be obtained via running EP including integrations on the graph
without variable $i$, we have a fast
algorithm for the responses, that does not involve integrations. Thus the
cost for a cavity covariance matrix is determined by the factorizing EP
updates on the graph without the central variable $i$, after which the
responses may quickly be obtained. 
This way of estimating $A_i$ is obviously as costly as a number of inversions
of matrices of dimension $N$ In the full Gaussian EP approach, an 
inversion is necessary at each update, such that in the end the present 
approach might become cheaper, but for the current model EP itself seems
more practicle. 
Possibly however, for more complex models the loop correction approach 
is beneficial, a topic which is to be further investigated.

\section{Discussion}
In this paper we have derived loop corrected belief propagation equations
for continuous variable models. In particular we have worked out the 
exactly tractable case of Gaussian models, and have derived the 
exact message passing equations. The role of the
``connected correlations'' of the cavity distribution discussed in 
\cite{Rizzo} is taken over by the off-diagonal parts of the covariance
matrix, which may be obtained in preprocessing algorithms. Moreover, 
using the fact that if ordinary BP converges, it produces exact 
marginal averages, various relations between BP messages are obtained,
leading to alternative update schemes to invert the covariance matrix.

For models involving nonlinear terms, for which approximation algorithms
are needed in order to compute marginals, we discussed some relations between
expectation propagation approaches and loop correction strategies, in 
particular for a model class where the nonlinear potentials involve only 
single variables. 
Loop correction approaches for continuous variable models 
become attractive once alternative strategies like 
expectation propagation grow too costly. On the other hand, however,
they are themselves heavily 
penalized by the cost of the preprocessing
stage necessary to estimate the cavity connected correlations.

\appendix
\section{Full covariance matrix as a function of BP quantities}
Directly writing out the error $\alpha_i^{\rm LC}-\alpha_i^{\rm BP}$
results in
\begin{eqnarray}
\alpha_i^{\rm LC}-\alpha_i^{\rm BP}& =& 
\sum_{k,l \in \partial i} J_{il}\left[\delta_{kl}[v_l^{i\ {\rm LC}}-v_l^{i
\ {\rm BP}}]+ (A_i)_{kl} (1-\delta_{kl})\right]J_{ik} 
\end{eqnarray}
It follows that
\begin{eqnarray}
J_{il}[v_l^{i\ {\rm LC}}-v_l^{i
\ {\rm BP}}]
+ \sum_{k \in \partial i \setminus l}J_{ik} (A_i)_{kl}
 =  -\frac{m_l^{(i) \ {\rm BP}}
- m_l^{i \ {\rm BP}}}{m_i^{\rm BP}}
\end{eqnarray}
from which we may furthermore derive the relation
\begin{eqnarray}
\left\{(D_i + A_i)\mathbf{J}_i\right\}_j = J_{ij}v_j^{i{\rm BP}}
-\frac{[m_j^{(i){\rm BP}}-m_j^{i\ {\rm BP}}]}{m_i^{\rm BP}}  
\label{funnyrelation}
\end{eqnarray}
\subsection{Off-diagonal parts of covariance matrix}
The off-diagonal elements of the covariance matrix of the model may be 
expressed in terms of cavity distributions again, using the formulation 
of subsection (\ref{basicideas}). We find, using the measures 
(\ref{measurei}),
\begin{eqnarray}
\langle \sigma_i \sigma_j \rangle & = & \mu_i \langle \sigma_j \rangle_i 
+ s_i \sum_{l \in \partial_i} J_{il} \langle \sigma_l \sigma_j \rangle_i
\nonumber \\
& = & m_i^{\rm BP} m_j^{(i)\ {\rm BP}} + [(m_i^{\rm BP})^2+v_i^{\rm LC}]
\left[J_{ij}v_j^{i\ {\rm BP}}
-\frac{[m_j^{(i){\rm BP}}-m_j^{i\ {\rm BP}}]}{m_i^{\rm BP}}\right] 
\label{nncov}
\end{eqnarray}
where we have used (\ref{funnyrelation}).

Next nearest neighbor correlations follow from a similar calculation as
above:
\begin{eqnarray}
\langle \sigma_j \sigma_k \rangle & =& \langle \sigma_j \sigma_k \rangle_i 
\qquad \qquad \qquad \qquad \qquad j,k \in \partial_i
\nonumber \\
&= & m_j^{(i) \ {\rm BP}} m_k^{(i)\ {\rm BP}} 
+ m_i^{\rm BP}\left[m_j^{(i)\ {\rm BP}} \left\{(D_i+A_i)J_i\right\}_k
+m_k^{(i)\ {\rm BP}}
\left\{(D_i+A_i)J_i\right\}_j\right] \nonumber \\
&&
+ [(m_i^{\rm BP})^2 + v_i^{\rm LC}]
\left\{(D_i+A_i)J_i\right\}_j
\left\{(D_i+A_i)J_i\right\}_k + (D_i+A_i)_{jk}
\end{eqnarray}
All terms follow from BP on the graph with and without $i$, except for 
the last one. Using (\ref{funnyrelation}), and renaming terms, we obtain
equations (\ref{growcov}).

\end{document}